\documentclass[a4paper]{cas-dc}

\usepackage[numbers]{natbib}
\usepackage{placeins}
\usepackage{tabularx}

\newcolumntype{Z}{>{\raggedright\arraybackslash}X}
\newcommand{\prompt}[1]{{\small\texttt{#1}}}

\renewcommand{\thesection}{\Roman{section}}
\renewcommand{\thesubsection}{\Alph{subsection}}

\makeatletter
\renewcommand{\p@subsubsection}{\thesection.\thesubsection.}
\renewcommand{\@seccntformat}[1]{%
  \ifstrequal{#1}{subsubsection}
    {\csname the#1\endcsname)\hskip0.5em}
    {\csname the#1\endcsname\@seccntDot\hskip0.5em}}
\makeatother

\ExplSyntaxOn
\cs_set:Npn \__make_tbl_caption:nn #1#2
{
  \l_tbl_align_tl
  \skip_vertical:N \l_tbl_abovecap_skip
  {
    \parbox{\dimexpr(\l_tbl_width_dim)}
    {
      \centering
      \sffamily\small
      \textbf{\color{scolor}#1}\par
      #2\par\vskip4pt
    }
  }
  \skip_vertical:N \l_tbl_belowcap_skip
}
\ExplSyntaxOff

\begin{document}
\let\WriteBookmarks\relax
\def\floatpagepagefraction{1}
\def\textpagefraction{.001}

\shorttitle{GeoDistill-Refine}
\shortauthors{Y. Zhang et~al.}

\title [mode = title]{GeoDistill-Refine: Silhouette-First Geometry Distillation for Annotation-Free Spacecraft Segmentation}

\author[1]{Yonglong Zhang}

\author[1]{Zongwu Xie}

\author[1]{Yang Liu}[orcid=0000-0002-0773-0706]
\cormark[1]
\ead{E-mail address:liuyanghit@hit.edu.cn}

\cortext[cor1]{Corresponding author}

\affiliation[1]{organization={School of Mechatronics Engineering, Harbin Institute of Technology},
  city={Harbin},
  state={Heilongjiang},
  country={China}}

\begin{abstract}
Foundation segmentation models can provide supervision for spacecraft imagery without manual training masks, but their predictions vary with textual prompts and may contain geometric errors that are amplified during distillation. This paper presents GeoDistill-Refine, a two-stage framework that transfers offline SAM 3 pseudo-masks to a compact segmentation network. Six fixed prompts are fused by an unweighted 50\% vote to stabilize the teacher output. The student first learns the foreground silhouette and is then refined with signed-distance-field, skeleton, and area objectives derived from the pseudo-mask. A sample-level gate, computed from prompt agreement, the valid-prompt ratio, and pseudo-mask area plausibility, reduces the influence of unreliable pseudo-geometry. On the SpaceSense-Bench HJM lockbox set, GeoDistill-Refine improves Image IoU and Boundary F1 by 0.0456 and 0.1380, respectively, over a plain pseudo-label student. External evaluations on the SPEED+ Lightbox and Sunlamp domains and on TANGO show competitive regional overlap together with gains in boundary quality or foreground precision. The deployed TinyUNet contains 0.263 M parameters and requires approximately 1.1 ms per image on an RTX 4090; SAM 3 pseudo-mask construction and the auxiliary geometry branches are used only during training.
\end{abstract}


\maketitle

\section{Introduction}\label{sec:introduction}

On-orbit servicing, active debris removal, and ren\-dez\-vous with non-cooperative targets require robust visual perception of spacecraft. Foreground segmentation provides pixel-level spatial constraints for target localization, relative pose estimation, and three-dimensional reconstruction \cite{pauly2023survey,dung2021spacecraft,muniyasamy2026segmentation}. Segmenting previously unseen spacecraft nevertheless remains difficult because illumination, background, apparent scale, and structural configuration vary substantially. Results reported on SPEED+, TANGO, and SpaceSense-Bench have shown the performance degradation associated with illumination-domain shifts and changes in target geometry \cite{park2022speedplus,bechini2023dataset,wu2026spacesense}.

Most existing approaches depend on manually annotated pixels. Such annotations are costly because real on-orbit imagery is scarce and spacecraft often contain low-contrast boundaries and slender appendages. Simulation provides inexpensive labels but cannot fully reproduce real materials, illumination, or background distributions. Large foundation models offer a different route to supervision, although their memory and computational requirements make direct onboard deployment impractical.

Open-vocabulary detectors and promptable segmentation models can generate object boxes or masks for unlabeled images from textual prompts \cite{liu2024groundingdino,kirillov2023sam,carion2025sam3}. Recent spacecraft studies have demonstrated automatic mask generation, pseudo-label distillation into compact students, and structured prompting \cite{zhang2024automatic,hicsonmez2026annotationfree,welsh2026postlaunch}. The resulting teacher predictions, however, remain sensitive to wording, target scale, and local structure. Existing pseudo-label processing commonly relies on augmentation consistency, box fusion, or confidence filtering, with limited treatment of boundary, connectivity, and area errors. Distance-field supervision can improve spacecraft boundary modeling \cite{velentzas2026gabi}, but dense geometric targets derived directly from noisy pseudo-masks may propagate and spatially amplify teacher errors.

This paper introduces GeoDistill-Refine, an annotation-free training framework for compact spacecraft segmentation. A frozen SAM 3 teacher first generates offline pseudo-masks through multi-prompt consensus. The student then learns the primary silhouette before being refined with reliability-gated pseudo-geometry from the corresponding checkpoint. Only the single-frame RGB student is retained at deployment. Here, annotation-free refers specifically to student optimization without manual masks; validation annotations are still used for checkpoint selection and threshold calibration.

SpaceSense-Bench is used to assess generalization to unseen spacecraft. SPEED+ and TANGO instead provide external validation of in-domain distillation under challenging illumination and different image distributions; they are not used to claim unseen-spacecraft generalization. On the held-out HJM lockbox set, GeoDistill-Refine improves Image IoU and Boundary F1 by 0.0456 and 0.1380, respectively, over a plain pseudo-label student. Ablation results further support the silhouette-first warm start and reliability-gated signed distance field (SDF) refinement under the evaluated student and loss configuration.

The main contributions are as follows:
\begin{enumerate}
  \item A fixed multi-prompt consensus teacher reduces the instability of single-prompt pseudo-masks and supplies more stable spacecraft foreground supervision without introducing the large teacher into deployment.
  \item A silhouette-first two-stage schedule separates initial foreground learning from low-weight geometric refinement. Under the evaluated TinyUNet configuration, it is more effective than optimizing noisy pseudo-geometry jointly from random initialization.
  \item A sample-level reliability gate, derived from prompt agreement, the valid-prompt ratio, and pseudo-mask area plausibility, modulates the SDF, skeleton, and area objectives. It reduces the contribution of anomalous pseudo-geometry while retaining the majority-voted mask supervision for every training sample.
\end{enumerate}

\section{Related Work}\label{sec:related-work}

\subsection{Spacecraft Segmentation and Generalization to Unseen Targets}\label{sec:spacecraft-segmentation}

Visual spacecraft perception encompasses target detection, component segmentation, and relative pose estimation. The original SPEED challenge established a common synthetic-to-real benchmark and documented the data scarcity and domain gap that distinguish spaceborne vision from terrestrial perception \cite{kisantal2020speed}. SPEED+ and TANGO subsequently provided synthetic or hardware-in-the-loop imagery for studying illumination changes and imaging-domain gaps \cite{park2022speedplus,bechini2023dataset}, while spacecraft-part datasets support detection and semantic segmentation benchmarks \cite{dung2021spacecraft}. SpaceSense-Bench broadens the range of spacecraft and sensing modalities and uses a spacecraft-disjoint protocol to emphasize the difficulty of unseen targets and small structures \cite{wu2026spacesense}.

Many segmentation systems combine encoder--decoder architectures with lightweight backbones to balance contour recovery and computational cost \cite{ronneberger2015unet,howard2019mobilenetv3,muniyasamy2026segmentation}. Most nevertheless require manual labels and focus on known targets or imaging-domain transfer. Adaptation to previously unseen spacecraft structures in the absence of training masks remains less thoroughly studied.

\subsection{Foundation-Model Pseudo-Labels and Annotation-Free Distillation}\label{sec:pseudo-label-distillation}

Pseudo-labeling, self-training, and knowledge distillation provide established mechanisms for exploiting unlabeled imagery \cite{lee2013pseudolabel,xie2020noisystudent,hinton2015distillation}. Mean Teacher and FixMatch stabilize supervision through temporal averaging or augmentation consistency \cite{tarvainen2017meanteacher,sohn2020fixmatch}; segmentation methods such as Cross Pseudo Supervision, ST++, and U2PL further use mutual prediction, checkpoint stability, or low-confidence evidence \cite{chen2021cps,yang2022stpp,wang2022u2pl}. These methods establish the importance of pseudo-label reliability, but generally remain semi-supervised and estimate quality from model confidence or training-time consistency.

Grounding DINO, SAM, SAM 3, and SEEM can produce boxes or masks directly from textual and visual prompts \cite{liu2024groundingdino,kirillov2023sam,carion2025sam3,zou2023seem}, reducing the need for a task-specific teacher. Their frozen predictions nevertheless vary with wording and prompt conditions. A confidence score for one prediction neither measures agreement across valid descriptions nor exposes errors shared by several prompts.

In spacecraft vision, foundation models have supported automatic annotation, lightweight student distillation, and prompt-driven capability extension \cite{zhang2024automatic,hicsonmez2026annotationfree,welsh2026postlaunch}. Existing studies emphasize annotation generation, candidate fusion, or iterative distillation; inter-prompt disagreement and the reliability of geometry derived from pseudo-masks remain less explored. GeoDistill-\allowbreak Refine uses fixed multi-prompt consensus while reserving its reliability estimate for pseudo-geometric supervision, separating prompt stabilization from the treatment of anomalous geometry.

\subsection{Boundary Modeling and Distance-Field Supervision}\label{sec:geometry-supervision}

Boundary-aware segmentation complements region objectives with explicit contour or topology cues. Gated-SCNN maintains a dedicated shape stream, distance-transform and boundary losses encode displacement from reference contours, and clDice targets the connectivity of thin structures \cite{takikawa2019gatedscnn,audebert2019distance,chai2020distance,kervadec2021boundary,shit2021cldice}. These approaches extend supervision beyond pixelwise overlap, but their geometric targets are typically derived from manual masks.

Related ideas have also been explored for spacecraft segmentation. GABI combines distance information with boundary features for contour recovery under difficult imaging conditions \cite{velentzas2026gabi}. Annotation-free distillation changes the underlying assumption because errors in a pseudo-mask propagate into its SDF, skeleton, and area targets. Applying such objectives without accounting for target reliability can therefore amplify teacher errors. GeoDistill-\allowbreak Refine addresses this gap by introducing pseudo-geometric supervision after silhouette learning and modulating it with a sample-level reliability gate.

\begin{figure*}[t]
  \centering
  \includegraphics[width=\textwidth]{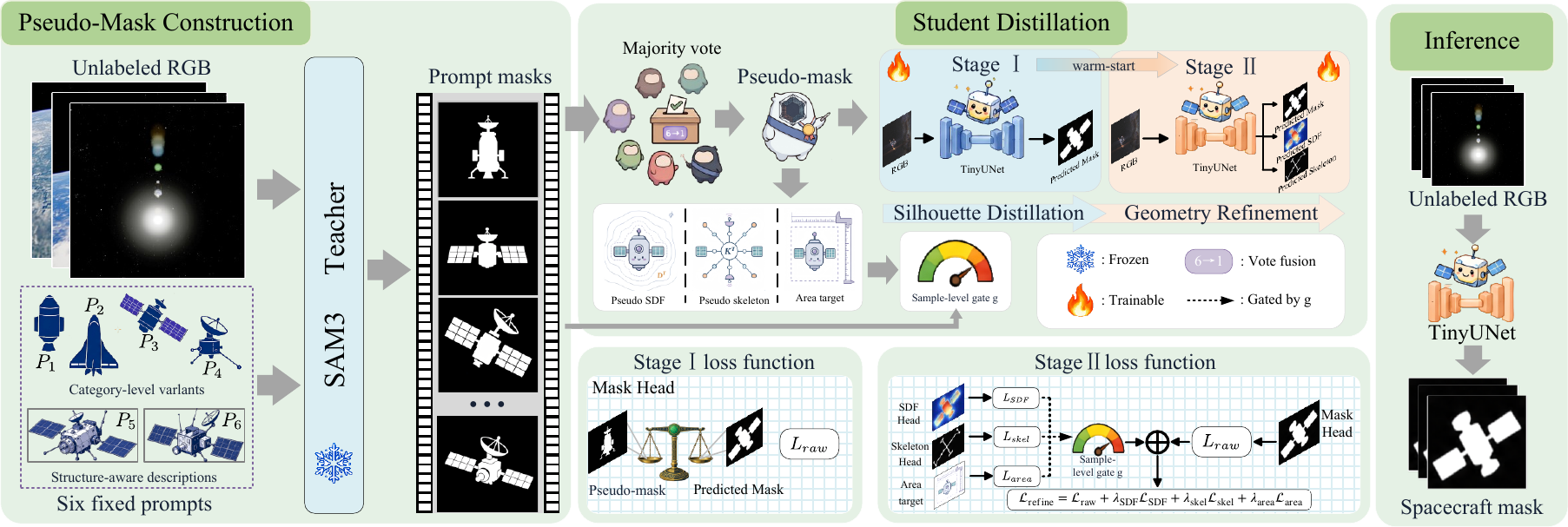}
  \caption{Overview of GeoDistill-Refine. Six fixed prompts are fused into a pseudo-mask that supplies the silhouette target, pseudo-geometric targets, and sample-level reliability gate. Stage II warm-starts the Stage-I student and adds gated geometry refinement; only the TinyUNet mask path is retained for inference.}
  \label{fig:overview}
\end{figure*}

\section{Method}\label{sec:method}

We consider single-frame binary spacecraft foreground segmentation under annotation-free student training. Each input image satisfies $I_i\in\mathbb{R}^{3\times H\times W}$, and the unlabeled training set is
\begin{equation}
\mathcal{D}_{u}=\{I_i\}_{i=1}^{N}
\label{eq:dataset}
\end{equation}
For the unseen-spacecraft evaluation on SpaceSense-Bench, the training and evaluation sets are separated by spacecraft identity. Let $\mathcal{S}_{\mathrm{tr}}$ and $\mathcal{S}_{\mathrm{ev}}$ denote their respective sets of spacecraft. This protocol requires
\begin{equation}
\mathcal{S}_{\mathrm{tr}}\cap\mathcal{S}_{\mathrm{ev}}=\varnothing
\label{eq:spacecraft-disjoint}
\end{equation}
The objective is to learn a student network $f_{\theta}$ that predicts the foreground logit $Z_i$ and probability map $P_i=\sigma(Z_i)$ from $I_i$. Annotation-free refers only to optimization of the student: manual masks are excluded from the training loss and are used solely for validation-time checkpoint selection, threshold calibration, and final evaluation.

As summarized in Fig.~\ref{fig:overview}, GeoDistill-Refine comprises offline pseudo-mask construction, two-stage student distillation, and student-only inference. A frozen SAM 3 teacher first generates six prompt-specific masks, whose majority vote forms the teacher pseudo-mask. Stage I trains the TinyUNet mask path to learn a binary silhouette. Stage II warm-starts the same student from the selected Stage-I checkpoint and adds pseudo-geometric supervision; the two TinyUNet icons in Fig.~\ref{fig:overview} therefore denote successive training stages of one student rather than separate deployed models. This is label distillation from hard teacher masks; no teacher features or logits are matched.

\subsection{Multi-Prompt Consensus-Based Pseudo-Mask Construction}\label{sec:multi-prompt}

\begin{figure*}[t]
  \centering
  \includegraphics[width=\textwidth]{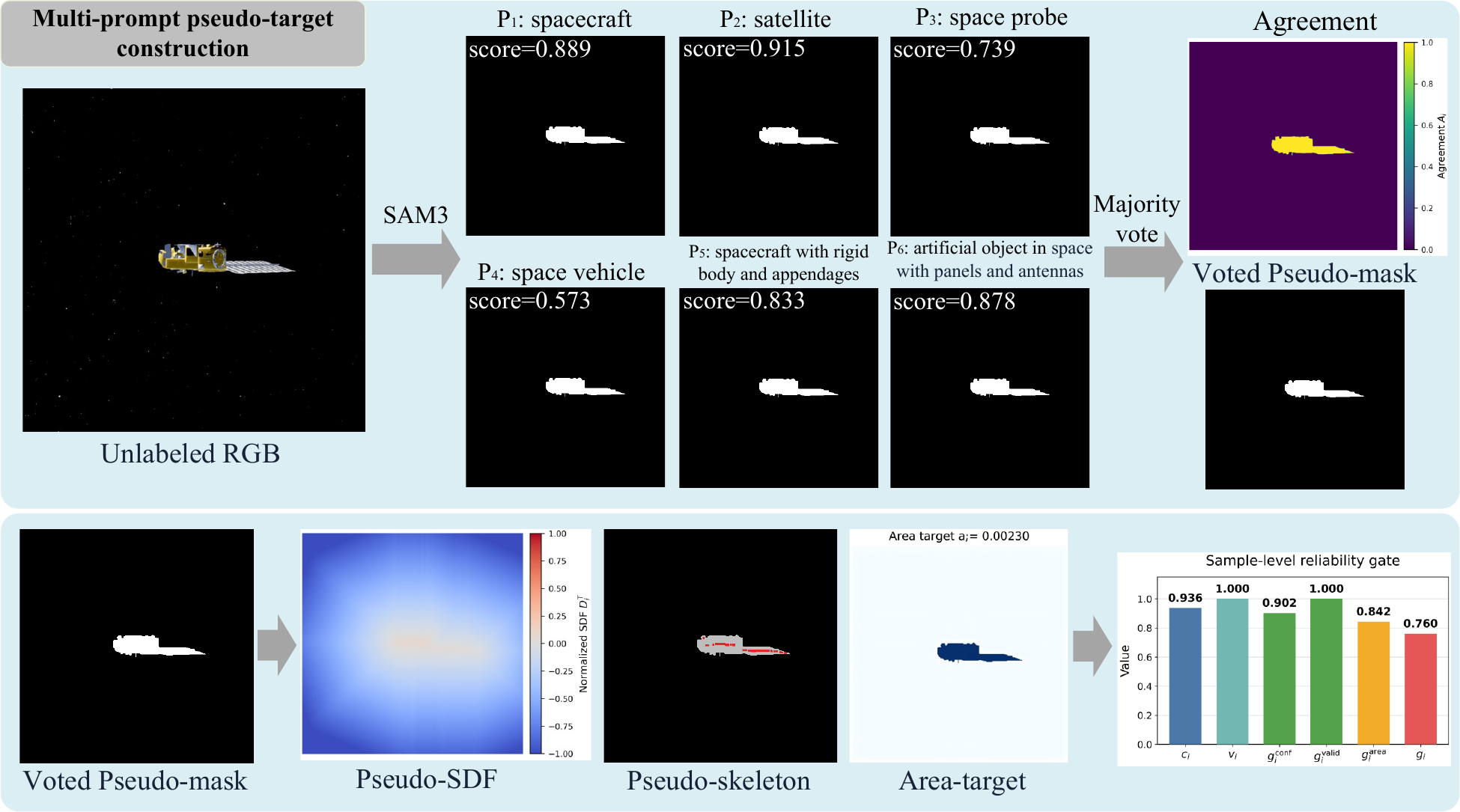}
  \caption{Multi-prompt pseudo-target construction for an unlabeled Landsat~8 training image. Six fixed prompts produce candidate masks and teacher scores. The valid masks form the agreement map $A_i$, whose 50\% consensus vote yields $M_i^T$. The normalized SDF $D_i^T$, skeleton $K_i^T$ , and area target $a_i$ are then derived from $M_i^T$. The sample-level gate combines foreground agreement $c_i$, valid-prompt ratio $v_i$, and area plausibility through Eqs.}
  \label{fig:pseudo-geometry}
\end{figure*}

\begingroup\setlength{\emergencystretch}{1em}
For each training image, the frozen SAM 3 teacher is queried with six fixed foreground prompts. Four are category-level descriptions---\allowbreak\prompt{spacecraft}, \prompt{satellite}, \prompt{space probe}, and \prompt{space vehicle}---and two add structural context: \prompt{spacecraft with rigid body and appendages} and \prompt{artificial object in space with panels and antennas}. Every prompt denotes the complete spacecraft foreground rather than an independent semantic component. The fixed prompt set is
\par\endgroup
\begin{equation}
\mathcal{P}=\{p_k\}_{k=1}^{6}
\label{eq:prompt-set}
\end{equation}
The graphical labels $P_1$--$P_6$ in Figs.~\ref{fig:overview} and~\ref{fig:pseudo-geometry} identify the six prompts $p_1$--$p_6$ in $\mathcal{P}$. In the equations, lowercase $p_k$ denotes a text prompt, whereas uppercase $P_i$ remains reserved for the student's foreground probability map.
For image $I_i$ and prompt $p_k\in\mathcal{P}$, the teacher $\mathcal{T}$ returns its highest-scoring candidate mask $M_{ik}\in\{0,1\}^{H\times W}$ and the associated score $s_{ik}$:
\begin{equation}
\left(M_{ik},s_{ik}\right)=\mathcal{T}(I_i,p_k)
\label{eq:teacher-prediction}
\end{equation}
Only predictions with $s_{ik}>0$ participate in fusion. Let $\mathcal{V}_i=\{k\mid s_{ik}>0\}$ be the valid-prompt set. For $|\mathcal{V}_i|>0$, the pixelwise agreement map is
\begin{equation}
A_i(x)=\frac{1}{|\mathcal{V}_i|}
\sum_{k\in\mathcal{V}_i}M_{ik}(x)
\label{eq:agreement-map}
\end{equation}
The teacher pseudo-mask is obtained by an unweighted 50\% consensus vote:
\begin{equation}
M_i^{T}(x)=
\mathbb{I}\!\left[A_i(x)\geq 0.5\right]
\label{eq:majority-vote}
\end{equation}
Thus, a tie among the valid prompts is retained as foreground; for six valid prompts, at least three foreground votes are required. We retain the conventional term ``majority vote'' in the experimental tables for brevity.
The scores $s_{ik}$ are not used to reweight individual pixels. If none of the six prompts yields a positive-score prediction, we set $A_i(x)=0$, $M_i^T(x)=0$, and the corresponding geometry gate to zero. Prompt-specific masks, scores, and agreement maps are generated once and cached offline; $M_i^T$ is reconstructed according to the rule above when the training sample is loaded.

\subsection{Compact Student Distillation for Stable Silhouette Learning}\label{sec:student-distillation}

The student is a TinyUNet operating at $384\times384$ resolution. Its three encoder blocks contain 24, 48, and 96 channels, and the decoder restores spatial resolution through two upsampling levels with skip connections. The mask, SDF, and skeleton heads share the decoded features and produce $P_i$, $D_i^S$, and $Z_i^K$, respectively. The area prediction is computed directly from $P_i$ and therefore introduces no additional head. Only the mask branch is active in Stage I; the SDF and skeleton heads are initialized at the start of Stage II.

Let $\mathcal{B}$ index a batch of size $B$. Foreground class imbalance is handled by a positive-class weight shared across the batch:
\begin{equation}
N_{+}
=\max\!\left(
\sum_{i\in\mathcal{B}}\sum_x M_i^T(x),1
\right)
\label{eq:positive-count}
\end{equation}
\begin{equation}
N_{-}
=\max\!\left(
\sum_{i\in\mathcal{B}}\sum_x [1-M_i^T(x)],1
\right)
\label{eq:negative-count}
\end{equation}
\begin{equation}
\pi=\operatorname{clip}\!\left(\frac{N_{-}}{N_{+}},1,500\right)
\label{eq:positive-weight}
\end{equation}
The resulting balanced binary cross-entropy is
\begin{equation}
\begin{aligned}
\mathcal{L}_{\mathrm{BBCE}}
={}&-\frac{1}{BHW}
\sum_{i\in\mathcal{B}}\sum_x
\left[
\pi M_i^T(x)\log P_i(x) \right.\\
&\left.\qquad
+\bigl(1-M_i^T(x)\bigr)\log\bigl(1-P_i(x)\bigr)
\right]
\end{aligned}
\label{eq:bbce}
\end{equation}
The Dice term is computed per image and then averaged over the batch:
\begin{equation}
\mathcal{L}_{\mathrm{Dice}}
=1-\frac{1}{B}\sum_{i\in\mathcal{B}}
\frac{2\sum_x P_i(x)M_i^T(x)+1}
{\sum_x P_i(x)+\sum_x M_i^T(x)+1}
\label{eq:dice}
\end{equation}
\begin{equation}
\mathcal{L}_{\mathrm{raw}}
=\mathcal{L}_{\mathrm{BBCE}}
+0.2\,\mathcal{L}_{\mathrm{Dice}}
\label{eq:raw-loss}
\end{equation}
Stage I is trained from random initialization for 20 epochs. The checkpoint selected on the validation set supplies the encoder, decoder, and mask-head parameters for Stage II.

\subsection{Reliability-Aware Pseudo-Geometry Refinement}\label{sec:geometry-refinement}

The SDF, skeleton, and area targets visualized in Fig.~\ref{fig:pseudo-geometry} are all derived from the voted pseudo-mask $M_i^T$. Imposing them jointly from random initialization would expose the student to both binary contour errors and their geometric transformations before a stable foreground has formed. Stage II is therefore not reinitialized. Instead, low-weight geometric refinement begins from the Stage-I checkpoint, with the SDF serving as the principal continuous geometric target.

Let $d_{\mathrm{in}}(x)$ and $d_{\mathrm{out}}(x)$ be the interior and exterior distance transforms of $M_i^T$, respectively, and define $\widetilde{D}_i^T=d_{\mathrm{in}}-d_{\mathrm{out}}$. For a non-degenerate mask, let $q_i=q_{0.95}(|\widetilde{D}_i^T|)$. We set $s_i=q_i$ when $q_i\geq10^{-6}$ and $s_i=1$ otherwise. The normalized target is
\begin{equation}
D_i^T(x)=
\begin{cases}
0, & \max_x M_i^T(x)=0,\\
1, & \min_x M_i^T(x)=1,\\
\operatorname{clip}\!\left(
\dfrac{\widetilde{D}_i^T(x)}{s_i},-1,1
\right), & \text{otherwise}
\end{cases}
\label{eq:sdf-target}
\end{equation}
The auxiliary skeleton target $K_i^T=\operatorname{Skel}(M_i^T)$ consists of local maxima in the interior distance map after removing points whose distance does not exceed one pixel. If a non-empty mask yields no valid skeleton, $K_i^T$ falls back to $M_i^T$. For visibility, Fig.~\ref{fig:pseudo-geometry} overlays this one-pixel skeleton in red on the voted pseudo-mask. The pseudo-foreground area ratio is
\begin{equation}
a_i=\frac{1}{HW}\sum_x M_i^T(x)
\label{eq:pseudo-area}
\end{equation}

For the Landsat~8 example in Fig.~\ref{fig:pseudo-geometry}, all six prompts are valid and the foreground agreement is $c_i=0.936$. Its pseudo-foreground area is $a_i=0.00230$, approximately $1.17$ times the training-set median $a_0=0.00197$. The resulting factors are $g_i^{\mathrm{conf}}=0.902$, $g_i^{\mathrm{valid}}=1.000$, and $g_i^{\mathrm{area}}=0.842$, giving a final reliability weight of $g_i=0.760$. Thus, mutually consistent prompt predictions with a plausible foreground area retain substantial pseudo-geometric supervision. More generally, the proxy combines mean agreement within the voted foreground $c_i$, the valid-prompt ratio $v_i$, and the pseudo-mask area ratio $a_i$:
\begin{equation}
c_i=
\begin{cases}
\dfrac{\sum_x A_i(x)M_i^T(x)}
{\sum_x M_i^T(x)}, & \sum_x M_i^T(x)>0,\\
0, & \text{otherwise}
\end{cases}
\label{eq:foreground-agreement}
\end{equation}
\begin{equation}
v_i=\frac{|\mathcal{V}_i|}{6}
\label{eq:valid-prompt-ratio}
\end{equation}
Let $a_0$ denote the median of $a_i$ over non-empty training pseudo-masks, and set $\varepsilon=10^{-6}$ for numerical stability. These signals are converted into confidence, valid-prompt, and area-plausibility gate factors.
\begin{equation}
g_i^{\mathrm{conf}}
=\operatorname{clip}\!\left(
\frac{c_i-0.35}{1-0.35},0,1
\right)
\label{eq:confidence-gate}
\end{equation}
\begin{equation}
g_i^{\mathrm{valid}}
=\operatorname{clip}\!\left(
\frac{v_i-0.20}{1-0.20},0,1
\right)
\label{eq:validity-gate}
\end{equation}
\begin{equation}
g_i^{\mathrm{area}}
=\operatorname{clip}\!\left[
\exp\!\left(
-\frac{\left|\log\frac{a_i+\varepsilon}{a_0+\varepsilon}\right|}{0.9}
\right),0.05,1
\right]
\label{eq:area-gate}
\end{equation}
\begin{equation}
g_i=\operatorname{clip}\!\left(
g_i^{\mathrm{conf}}\,
g_i^{\mathrm{valid}}\,
g_i^{\mathrm{area}},0,1
\right)
\label{eq:reliability-gate}
\end{equation}
The gate $g_i$ is applied only to the SDF, skeleton, and area terms, not to the primary segmentation loss. The dashed paths marked ``Gated by $g$'' in Fig.~\ref{fig:overview} denote this selective samplewise modulation; the equations use $g_i$ to make the sample index explicit. The gate thresholds were specified without KCP annotations and then fixed for all subsequent evaluations.

The SDF head produces $D_i^S\in[-1,1]^{H\times W}$ using a tanh activation. Distance errors near the pseudo-boundary receive greater weight:
\begin{equation}
B_i(x)
=\mathbb{I}\!\left[|D_i^T(x)|\leq0.55\right]
\label{eq:pseudo-boundary-band}
\end{equation}
\begin{equation}
w_i^{\mathrm{SDF}}(x)
=g_i\left[0.15+0.85B_i(x)\right]
\label{eq:sdf-weight}
\end{equation}
\begin{equation}
\mathcal{L}_{\mathrm{SDF}}
=
\frac{\sum_{i\in\mathcal{B}}\sum_x
w_i^{\mathrm{SDF}}(x)
\left|D_i^S(x)-D_i^T(x)\right|}
{\sum_{i\in\mathcal{B}}\sum_x
w_i^{\mathrm{SDF}}(x)+\varepsilon}
\label{eq:sdf-loss}
\end{equation}
Let $Z_i^K$ denote the skeleton-head logits. Reusing the batch-level balanced binary cross-entropy operator gives
\begin{equation}
\mathcal{L}_{\mathrm{skel}}
=\operatorname{BBCE}\!\left(
\{Z_i^K\}_{i\in\mathcal{B}},
\{K_i^T\}_{i\in\mathcal{B}};\{g_i\}_{i\in\mathcal{B}}
\right)
\label{eq:skeleton-loss}
\end{equation}
Here, $K_i^T$ is the target, and $g_i$ is broadcast as a per-pixel weight within each sample. The positive-class weight is computed from the gated counts of positive and negative skeleton pixels.

The area term introduces no separate prediction head. It compares the foreground proportions of the mask probability and the pseudo-mask through a logarithmic ratio error:
\begin{equation}
\widehat{a}_i=\frac{1}{HW}\sum_x P_i(x)
\label{eq:predicted-area}
\end{equation}
\begin{equation}
\ell_{\mathrm{area}}^{(i)}
=
\left|
\log\frac{\widehat{a}_i+10^{-5}}{a_i+10^{-5}}
\right|
\label{eq:area-error}
\end{equation}
\begin{equation}
\mathcal{L}_{\mathrm{area}}
=
\frac{\sum_{i\in\mathcal{B}} g_i\ell_{\mathrm{area}}^{(i)}}
{\sum_{i\in\mathcal{B}} g_i+\varepsilon}
\label{eq:area-loss}
\end{equation}
Stage II is warm-started from the Stage-I checkpoint and trained for 10 epochs. The SDF remains the principal geometric objective, while the skeleton and area terms act only as low-weight auxiliaries. Let $\lambda_{\mathrm{SDF}}$, $\lambda_{\mathrm{skel}}$, and $\lambda_{\mathrm{area}}$ denote the three coefficients displayed schematically in Fig.~\ref{fig:overview}. Because $g_i$ is already incorporated samplewise into Eqs.~\eqref{eq:sdf-loss}, \eqref{eq:skeleton-loss}, and \eqref{eq:area-loss}, it is not multiplied a second time after batch aggregation. The primary mask supervision remains ungated:
\begin{equation}
\mathcal{L}_{\mathrm{refine}}
=\mathcal{L}_{\mathrm{raw}}
+\lambda_{\mathrm{SDF}}\mathcal{L}_{\mathrm{SDF}}
+\lambda_{\mathrm{skel}}\mathcal{L}_{\mathrm{skel}}
+\lambda_{\mathrm{area}}\mathcal{L}_{\mathrm{area}}
\label{eq:refine-loss}
\end{equation}
The coefficients are fixed as $\lambda_{\mathrm{SDF}}=0.002$, $\lambda_{\mathrm{skel}}=0.001$, and $\lambda_{\mathrm{area}}=0.05$. For the default SpaceSense-Bench configuration, the final model is the Stage-II epoch-10 checkpoint. The quality-adaptive configuration used only for the external evaluations is specified in Section~\ref{sec:external-validation}. At inference, an RGB image passes only through the TinyUNet mask path. A threshold $\tau$ fixed on the validation set converts its probability map into the binary spacecraft mask:
\begin{equation}
\widehat{M}_i(x)=\mathbb{I}\!\left[P_i(x)\geq\tau\right]
\label{eq:inference-mask}
\end{equation}
Consequently, SAM 3, the prompt cache, the SDF head, and the skeleton head are absent from deployment; the area term has no prediction head to retain.

\section{Experiments}\label{sec:experiments}

\subsection{Experimental Setup}\label{sec:experimental-setup}

\subsubsection{Datasets}\label{sec:datasets}

\begingroup\setlength{\emergencystretch}{1em}
Experiments are conducted on SpaceSense-\allowbreak Bench, SPEED+, and TANGO. They respectively examine generalization to unseen spacecraft structures, the stability of pseudo-label distillation under challenging illumination, and validation on an independent dataset. All three datasets are converted to binary spacecraft foreground segmentation, and images are resized to $384\times384$.
\par\endgroup

\textbf{SpaceSense-Bench.} Only single-frame RGB images are used, and all non-background classes in the original semantic masks are merged into a single spacecraft foreground. The data are divided by spacecraft identity, with no target overlap across subsets. The training set contains 270 images of nine spacecraft, and the validation set contains 60 images of three spacecraft. Evaluation is performed on three additional unseen-spacecraft groups, KCP, CVI, and HJM, each comprising 60 images of three spacecraft. KCP and CVI provide development evidence on two different target combinations, with KCP also used for ablation. HJM is reserved as a lockbox set after the method has been frozen.

\textbf{SPEED+.} The Lightbox and Sunlamp hardware-in-the-loop domains are used to evaluate segmentation under diffuse and strong direct illumination. For each domain, 500 fixed images form the unlabeled training set, and non-overlapping validation and test subsets are drawn from the remaining images. Separate students and thresholds are used for the two domains. These experiments assess in-domain pseudo-label distillation under difficult illumination; they do not support claims of cross-spacecraft or cross-domain generalization.

\textbf{TANGO.} A fixed set of 500 images from the official training split is used as unlabeled training data, and a disjoint set of 500 images is used for validation. The official test set of 3,002 images is left unchanged and used only for final evaluation. TANGO tests whether similar behavior is observed under an independent acquisition setup and image distribution.

\begingroup\setlength{\emergencystretch}{1em}
The strict Annotation-Free reproduction retains its original teacher and pseudo-label processing pipeline. GeoDistill-\allowbreak Refine and its controlled comparisons update the student using only RGB images, SAM 3 voted pseudo-masks, and the corresponding pseudo-label processing outputs. No annotation-free method uses manual masks for student updates; manual segmentation masks are restricted to validation-time model selection, threshold calibration, and final evaluation.
\par\endgroup

\subsubsection{Training Details}\label{sec:training}

The frozen SAM 3 teacher generates six prompt-specific predictions in advance, and their 50\% consensus vote serves as the foreground pseudo-label. Plain Student and GeoDistill-Refine share the same TinyUNet and preprocessing. In the default SpaceSense-Bench protocol, Plain Student undergoes Stage-I silhouette learning only, whereas GeoDistill-Refine continues from the corresponding Stage-I checkpoint for Stage-II geometric refinement.

Both stages use AdamW with a batch size of 4 and weight decay of $10^{-4}$. For the default SpaceSense-Bench protocol, Stage I is trained from random initialization for 20 epochs with an initial learning rate of $5\times10^{-4}$, and Stage II runs for another 10 epochs with an initial learning rate of $10^{-4}$. The quality-adaptive external protocol and its checkpoint candidates are described in Section~\ref{sec:external-validation}. Main results use random seeds 42, 3407, and 2026; mechanistic ablations use seed 42 unless stated otherwise.

The binarization threshold is selected by Image IoU on the corresponding validation set and then applied unchanged to the test set. All experiments use PyTorch 2.5.1 and an NVIDIA GeForce RTX 4090 GPU.

\subsubsection{Metrics and Statistical Analysis}\label{sec:metrics}

The primary metrics are mean image-level intersection over union (Image IoU) and Boundary F1. Image IoU measures regional overlap between the predicted and reference foregrounds. Boundary F1 uses a two-pixel tolerance at $384\times384$ resolution and evaluates boundary displacement, contour gaps, and segmentation of slender structures.

Foreground precision and recall are also reported to diagnose over-segmentation and missed foreground. Predicted foreground area ratio and Small-region IoU are used as supplementary diagnostics; the latter is computed only for SpaceSense-Bench.

Main student results are reported as the mean and standard deviation over three random seeds. For the SpaceSense-Bench controlled comparisons in Table~\ref{tab:spacesense-results}, $95\%$ confidence intervals are estimated from matched per-image outputs using $10\,000$ paired hierarchical bootstrap resamples. External reproduced baselines for which only run-level summary statistics are available are compared descriptively through their means and standard deviations; no paired significance test is claimed. The primary Plain Student versus GeoDistill-Refine comparison on KCP is additionally evaluated at a fixed threshold of 0.5 to assess sensitivity to threshold calibration.

\begin{table}[H]
  \centering
  \caption{Pseudo-mask quality of different teacher prompting configurations on SpaceSense-Bench KCP.}
  \label{tab:teacher-prompt-quality}
  \scriptsize
  \setlength{\tabcolsep}{1.5pt}
  \begin{tabularx}{\columnwidth}{@{}Zcccc@{}}
    \toprule
    Teacher configuration & \makecell{Image\\IoU} & \makecell{Boundary\\F1} & Precision & Recall \\
    \midrule
    Base prompt: \texttt{spacecraft} & 0.4743 & 0.5374 & 0.5137 & 0.5069 \\
    Validation-selected single prompt & 0.5691 & 0.6456 & 0.6243 & 0.6065 \\
    Mean over six single prompts & \makecell{$0.3462$\\$\pm0.2423$} & \makecell{$0.3934$\\$\pm0.2745$} & \makecell{$0.3726$\\$\pm0.2625$} & \makecell{$0.3734$\\$\pm0.2602$} \\
    Six-prompt majority vote & \textbf{0.6833} & \textbf{0.7794} & \textbf{0.7375} & \textbf{0.7403} \\
    \bottomrule
  \end{tabularx}
\end{table}

\subsection{Multi-Prompt Teacher and Pseudo-Label Quality Analysis}\label{sec:teacher-analysis}

\subsubsection{Effectiveness of Multi-Prompt Fusion}\label{sec:prompt-fusion}

\begingroup\setlength{\emergencystretch}{1em}
The six fixed prompts are first evaluated on the SpaceSense-\allowbreak Bench validation set. The KCP development set is then used to compare the base prompt \prompt{spacecraft}, the validation-selected prompt, the mean across independent single-prompt evaluations, and six-prompt majority voting. The validation-selected prompt is \prompt{spacecraft with rigid body and appendages}. KCP annotations are not involved in prompt selection, and no student is trained in this analysis.
\par\endgroup

The ``mean over six single prompts'' in Table~\ref{tab:teacher-prompt-quality} is the arithmetic mean and sample standard deviation of six independently evaluated single-prompt configurations ($n=6$); it is not a mask-fusion result. The large dispersion shows sensitivity to wording. Relative to the validation-selected prompt, majority voting increases Image IoU and Boundary F1 by 0.1142 and 0.1338, respectively, a result consistent with complementary predictions across prompts. This voting rule is fixed for all subsequent experiments. Results for the remaining prompts and fusion variants are deferred to the supplementary material, and no claim of universal optimality is made.

\begin{figure*}[t]
  \centering
  \includegraphics[width=\textwidth]{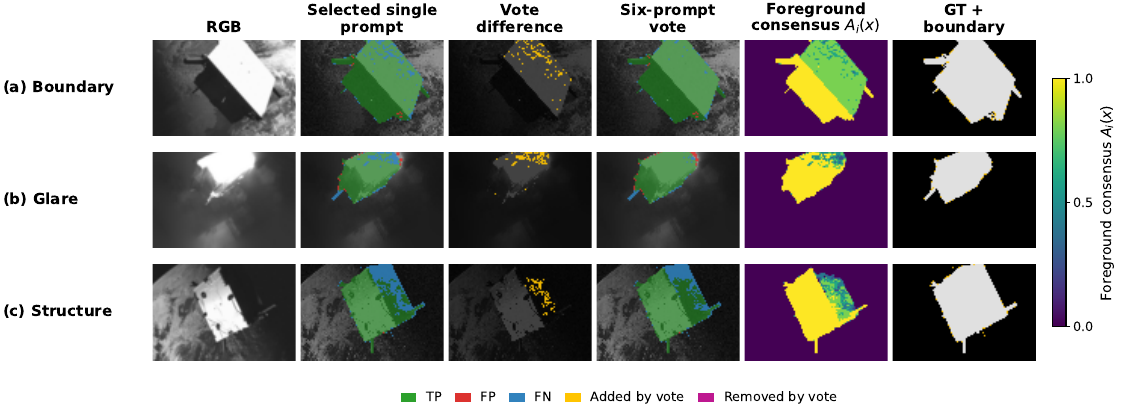}
  \caption{Qualitative analysis of multi-prompt pseudo-mask construction on full-frame SPEED+ images. The validation-selected single prompt is fixed for all cases. Green, red, and blue indicate true positives, false positives, and false negatives; yellow and magenta in the vote-difference maps indicate foreground added and removed by voting. $A_i(x)$ visualizes foreground consensus, and yellow contours mark ground-truth boundaries. Rows (a)--(c) show boundary completion, glare recovery, and structure recovery. Ground truth is used only for post-hoc visualization.}
  \label{fig:prompt-comparison}
\end{figure*}

Figure~\ref{fig:prompt-comparison} uses examples in which the spacecraft naturally occupies 33.8\%, 15.8\%, and 26.9\% of the full image, avoiding apparent enlargement from region-of-interest cropping. Voting raises pseudo-mask IoU from 0.920 to 0.966, from 0.838 to 0.917, and from 0.772 to 0.816 in rows (a)--(c), respectively. The vote-difference maps localize the added foreground responsible for these gains, while the intermediate $A_i(x)$ values identify boundaries or structures on which the valid prompts disagree. Thus, voting reduces prompt-specific omissions without implying that foreground consensus is a calibrated correctness estimate; prompt-consistent teacher errors can still remain.

\begin{table}[htbp]
  \centering
  \caption{Multi-prompt teacher pseudo-mask quality across datasets and imaging domains.}
  \label{tab:teacher-domain-quality}
  \scriptsize
  \setlength{\tabcolsep}{2.5pt}
  \begin{tabularx}{\columnwidth}{@{}Zcccc@{}}
    \toprule
    Dataset/domain & \makecell{Image\\IoU} & \makecell{Boundary\\F1} & Precision & Recall \\
    \midrule
    SpaceSense-Bench (KCP) & 0.6833 & 0.7794 & 0.7375 & 0.7403 \\
    SPEED+ Lightbox & 0.7721 & 0.7600 & 0.8823 & 0.7925 \\
    SPEED+ Sunlamp & 0.6235 & 0.5892 & 0.6725 & 0.6508 \\
    TANGO & 0.3324 & 0.3768 & 0.3858 & 0.3468 \\
    \bottomrule
  \end{tabularx}
\end{table}

\subsubsection{Pseudo-Label Quality and Reliability Analysis}\label{sec:reliability-analysis}

Multi-prompt voting does not remove errors caused by target scale, slender structures, or difficult illumination. Using the same prompt configuration, Table~\ref{tab:teacher-domain-quality} shows strong overlap and precision on Lightbox, degradation on Sunlamp, and both false positives and missed foreground on TANGO. KCP has lower Image IoU than Lightbox but the highest Boundary F1. These within-domain measurements show that regional and boundary noise need not vary together, making uniform pseudo-geometric supervision questionable.

The gate $g_i$ is analyzed post hoc on KCP using manual masks only to measure teacher quality; its inputs and parameters do not use KCP annotations. Its Spearman correlations with teacher Image IoU and Boundary F1 are $0.4230$ ($p=7.58\times10^{-4}$) and $0.5305$ ($p=1.30\times10^{-5}$). Because 18 samples have $g_i=0$ and the remaining 42 occupy a high-quality range, Table~\ref{tab:gate-groups} reports the actual gating states rather than equal-sized tertiles.
\begin{table}[htbp]
  \centering
  \caption{Teacher pseudo-mask quality for samples suppressed or retained by the reliability gate.}
  \label{tab:gate-groups}
  \scriptsize
  \setlength{\tabcolsep}{3pt}
  \begin{tabularx}{\columnwidth}{@{}Zccc@{}}
    \toprule
    Gating state & Images & \makecell{Teacher\\Image IoU} & \makecell{Teacher\\Boundary F1} \\
    \midrule
    $g_i=0$: geometry suppressed & 18 & 0.2853 & 0.3258 \\
    $g_i>0$: geometry enabled & 42 & 0.8539 & 0.9738 \\
    \bottomrule
  \end{tabularx}
\end{table}

The suppressed group has substantially lower teacher quality, indicating that the gate identifies anomalous pseudo-labels rather than providing a calibrated continuous quality estimate. This KCP analysis is mechanistic evidence only. During training, $g_i$ modulates the pseudo-geometric losses, while all voted masks remain in the primary segmentation loss, motivating the refinement evaluated next.

\subsection{Student Distillation and Multi-Dataset Evaluation}\label{sec:student-results}

\subsubsection{Results on Unseen Spacecraft in SpaceSense-Bench}\label{sec:unseen-results}

\begingroup\setlength{\emergencystretch}{1em}
Table~\ref{tab:spacesense-results} compares Plain Student, two pseudo-label processing baselines, a GABI-\allowbreak inspired pseudo-\allowbreak geometry baseline, and GeoDistill-Refine under a common protocol. All controlled methods share TinyUNet, teacher predictions, and unlabeled training images, except that the GABI-inspired baseline follows its joint-training formulation and is optimized from scratch. Results are means and standard deviations over three random seeds, and gains are calculated relative to Plain Student.
\par\endgroup

GeoDistill-Refine improves both Image IoU and Boundary F1 on KCP, CVI, and HJM. Relative to Plain Student, the gains are 0.0576 and 0.1017 on KCP and 0.0630 and 0.1376 on CVI. These two development sets show similar behavior for different combinations of unseen targets.

On the HJM lockbox set, Image IoU and Boundary F1 increase by 0.0456 and 0.1380, with corresponding $95\%$ confidence intervals of $[0.0107,\,0.0835]$ and $[0.0422,\,0.2496]$. Relative to the GABI-inspired baseline, GeoDistill-Refine gains 0.0302, 0.0330, and 0.0247 Image IoU across the three evaluation groups, together with 0.0568, 0.0588, and 0.0605 Boundary F1. The differences from this baseline were not subjected to separate significance tests.

\begin{table*}[t]
  \centering
  \caption{Segmentation results on the SpaceSense-Bench KCP and CVI development sets and HJM lockbox set.}
  \label{tab:spacesense-results}
  \scriptsize
  \begin{tabularx}{\textwidth}{cZccc}
    \toprule
    Evaluation set & Method & Image IoU & Boundary F1 & Image IoU gain and $95\%$ CI \\
    \midrule
    \multirow{5}{*}{KCP}
      & Plain Student & $0.5235\pm0.0458$ & $0.7173\pm0.0974$ & -- \\
      & Confidence-Filtered Student & $0.5022\pm0.0638$ & $0.6920\pm0.0826$ & $-0.0213\ [-0.0805,\,0.0426]$ \\
      & Annotation-Free-style Control & $0.5118\pm0.0564$ & $0.7056\pm0.0792$ & $-0.0117\ [-0.0612,\,0.0385]$ \\
      & GABI-inspired Pseudo-Geometry & $0.5509\pm0.0199$ & $0.7621\pm0.0151$ & $+0.0274\ [-0.0177,\,0.0831]$ \\
      & GeoDistill-Refine & $\mathbf{0.5811\pm0.0290}$ & $\mathbf{0.8189\pm0.0534}$ & $+0.0576\ [0.0351,\,0.0813]$ \\
    \midrule
    \multirow{5}{*}{CVI}
      & Plain Student & $0.3442\pm0.0607$ & $0.6013\pm0.1054$ & -- \\
      & Confidence-Filtered Student & $0.3496\pm0.0617$ & $0.6133\pm0.0927$ & $+0.0055\ [-0.0374,\,0.0448]$ \\
      & Annotation-Free-style Control & $0.3568\pm0.0561$ & $0.6237\pm0.0869$ & $+0.0126\ [-0.0294,\,0.0551]$ \\
      & GABI-inspired Pseudo-Geometry & $0.3743\pm0.0179$ & $0.6801\pm0.0435$ & $+0.0302\ [-0.0302,\,0.1113]$ \\
      & GeoDistill-Refine & $\mathbf{0.4073\pm0.0422}$ & $\mathbf{0.7389\pm0.0588}$ & $+0.0630\ [0.0166,\,0.1140]$ \\
    \midrule
    \multirow{5}{*}{HJM}
      & Plain Student & $0.3437\pm0.0570$ & $0.4689\pm0.1358$ & -- \\
      & Confidence-Filtered Student & $0.3479\pm0.0603$ & $0.4970\pm0.1291$ & $+0.0042\ [-0.0273,\,0.0315]$ \\
      & Annotation-Free-style Control & $0.3534\pm0.0548$ & $0.5127\pm0.1113$ & $+0.0097\ [-0.0236,\,0.0430]$ \\
      & GABI-inspired Pseudo-Geometry & $0.3646\pm0.0332$ & $0.5464\pm0.0586$ & $+0.0209\ [-0.0384,\,0.0928]$ \\
      & GeoDistill-Refine & $\mathbf{0.3893\pm0.0280}$ & $\mathbf{0.6069\pm0.0555}$ & $+0.0456\ [0.0107,\,0.0835]$ \\
    \bottomrule
  \end{tabularx}
\end{table*}

\begin{table*}[t]
  \centering
  \caption{Student segmentation results on SPEED+ Lightbox, SPEED+ Sunlamp, and TANGO.}
  \label{tab:external-results}
  \scriptsize
  \begin{tabularx}{\textwidth}{cZcccc}
    \toprule
    Data domain & Method & Image IoU & Boundary F1 & Precision & Recall \\
    \midrule
    \multirow{6}{*}{Lightbox}
      & Annotation-Free Reproduction & $\mathbf{0.6035\pm0.0148}$ & $0.4587\pm0.0215$ & $0.7984\pm0.0173$ & $0.7368\pm0.0196$ \\
      & Plain Student & $0.5419\pm0.0052$ & $0.4089\pm0.0472$ & $0.7482\pm0.0156$ & $0.6890\pm0.0133$ \\
      & Confidence-Filtered Student & $0.5554\pm0.0125$ & $0.3734\pm0.0359$ & $0.7070\pm0.0143$ & $0.7362\pm0.0048$ \\
      & Annotation-Free-style Control & $0.5648\pm0.0136$ & $0.3839\pm0.0314$ & $0.7188\pm0.0179$ & $\mathbf{0.7415\pm0.0112}$ \\
      & GABI-inspired Pseudo-Geometry & $0.5671\pm0.0174$ & $0.4071\pm0.0116$ & $0.7444\pm0.0135$ & $0.7233\pm0.0346$ \\
      & GeoDistill-Refine & $0.5983\pm0.0082$ & $\mathbf{0.4813\pm0.0126}$ & $\mathbf{0.8294\pm0.0065}$ & $0.7237\pm0.0188$ \\
    \midrule
    \multirow{6}{*}{Sunlamp}
      & Annotation-Free Reproduction & $\mathbf{0.6891\pm0.0127}$ & $0.5628\pm0.0184$ & $0.8059\pm0.0131$ & $\mathbf{0.8382\pm0.0167}$ \\
      & Plain Student & $0.6426\pm0.0121$ & $0.5169\pm0.0235$ & $0.7662\pm0.0096$ & $0.7894\pm0.0091$ \\
      & Confidence-Filtered Student & $0.6330\pm0.0237$ & $0.5178\pm0.0144$ & $0.7617\pm0.0183$ & $0.7784\pm0.0234$ \\
      & Annotation-Free-style Control & $0.6417\pm0.0204$ & $0.5226\pm0.0188$ & $0.7664\pm0.0162$ & $0.7879\pm0.0215$ \\
      & GABI-inspired Pseudo-Geometry & $0.6556\pm0.0091$ & $0.5323\pm0.0140$ & $0.7771\pm0.0158$ & $0.7997\pm0.0052$ \\
      & GeoDistill-Refine & $0.6823\pm0.0112$ & $\mathbf{0.5762\pm0.0111}$ & $\mathbf{0.8148\pm0.0086}$ & $0.8354\pm0.0172$ \\
    \midrule
    \multirow{6}{*}{TANGO}
      & Annotation-Free Reproduction & $\mathbf{0.8661\pm0.0071}$ & $\mathbf{0.9190\pm0.0063}$ & $0.9372\pm0.0126$ & $0.9254\pm0.0118$ \\
      & Plain Student & $0.7570\pm0.0343$ & $0.8703\pm0.0298$ & $0.8491\pm0.0527$ & $0.8975\pm0.0868$ \\
      & Confidence-Filtered Student & $0.8537\pm0.0068$ & $0.8973\pm0.0049$ & $0.9261\pm0.0207$ & $0.9178\pm0.0173$ \\
      & Annotation-Free-style Control & $0.8584\pm0.0079$ & $0.9011\pm0.0068$ & $0.9297\pm0.0191$ & $0.9208\pm0.0157$ \\
      & GABI-inspired Pseudo-Geometry & $0.7954\pm0.0531$ & $0.8979\pm0.0087$ & $0.9105\pm0.0211$ & $0.8721\pm0.0537$ \\
      & GeoDistill-Refine & $0.8642\pm0.0064$ & $0.9126\pm0.0115$ & $\mathbf{0.9618\pm0.0018}$ & $\mathbf{0.9282\pm0.0807}$ \\
    \bottomrule
  \end{tabularx}
\end{table*}

\subsubsection{External Validation on SPEED+ and TANGO}\label{sec:external-validation}

SPEED+ Lightbox, SPEED+ Sunlamp, and TANGO test in-domain distillation stability under different laboratory imaging conditions and consistency on an independent dataset. They do not evaluate generalization to unseen spacecraft identities. Pseudo-labels, students, checkpoints, and thresholds are produced independently for each domain.

For these external domains, we evaluate a pre-specified quality-adaptive extension rather than applying the fixed SpaceSense-Bench schedule unchanged. The proportions of pseudo-labels passing the confidence criterion in the Lightbox, Sunlamp, and TANGO training sets are 0.724, 0.544, and 0.308, respectively. Weak region anchoring to the frozen Stage-I prediction is retained for the first two domains. For TANGO, the Stage-II pseudo-mask region loss is disabled, and the unrefined Stage-I checkpoint, denoted as Stage-II epoch 0, remains eligible for model selection. This adaptation changes the region supervision and checkpoint candidate set but retains the frozen teacher, pseudo-geometric targets, and reliability gate. Its operating mode is determined only from training pseudo-label statistics using a fixed boundary of 0.5. Table~\ref{tab:external-results} reports three-seed results; test annotations are not used for training or model selection.

\begingroup\setlength{\emergencystretch}{1.5em}
Relative to the external reproduction baseline, GeoDistill-\allowbreak Refine is lower by 0.0052 and 0.0068 Image IoU on Lightbox and Sunlamp, respectively, while improving Boundary F1 by 0.0226 and 0.0134 and precision by 0.0310 and 0.0089. On TANGO, Image IoU and Boundary F1 are lower by 0.0019 and 0.0064, whereas precision and recall increase by 0.0246 and 0.0028. These are descriptive differences between means and are not interpreted as statistically significant.
\par\endgroup

Among methods using the same SAM 3 teacher and TinyUNet protocol, the quality-adaptive GeoDistill-Refine configuration achieves the highest mean Image IoU, Boundary F1, and precision in all three domains. The complete external configuration yields its clearest descriptive gains in boundary quality and false-positive control; the experiments do not isolate the effect of region anchoring alone.

\begin{figure*}[t]
  \centering
  \includegraphics[width=\textwidth]{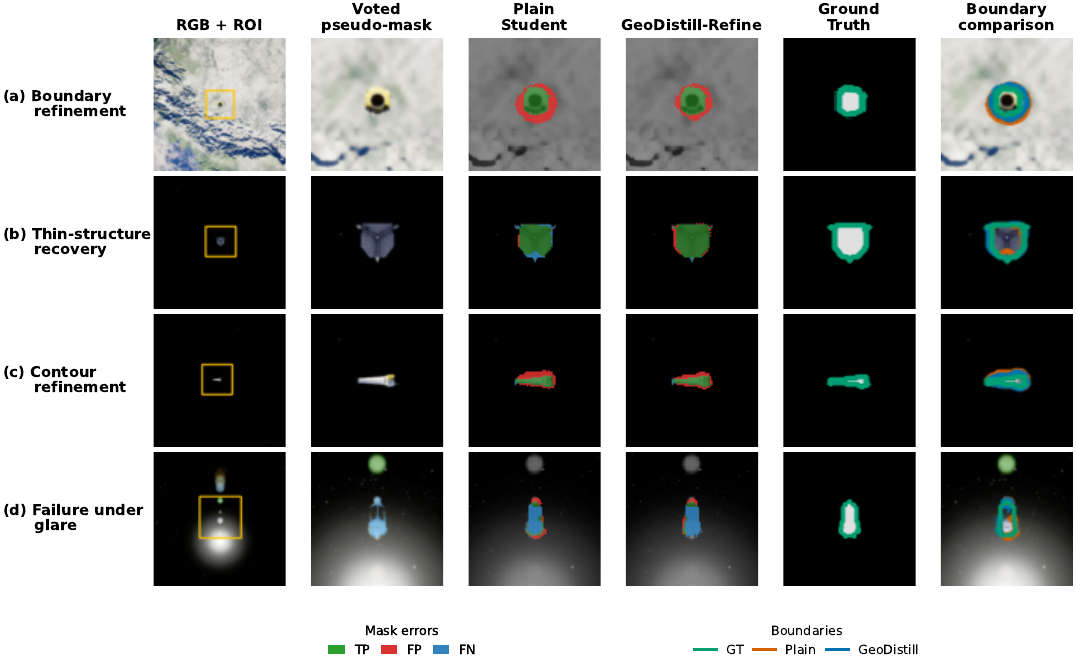}
  \caption{Qualitative comparison of Plain Student and GeoDistill-Refine on KCP using the seed-42 checkpoints and validation-calibrated thresholds. Yellow boxes identify identical visualization crops. Green, red, and blue in the student panels indicate true positives, false positives, and false negatives; green, orange, and blue contours denote ground truth, Plain Student, and GeoDistill-Refine. Rows (a)--(c) show boundary refinement on Kepler, thin-structure recovery on Philae, and contour refinement on Chandra; row (d) shows a Kepler failure under severe glare. All metrics are computed on full frames.}
  \label{fig:qualitative-results}
\end{figure*}

\subsubsection{Robustness, Efficiency, and Qualitative Results}\label{sec:robustness-efficiency}

At the fixed KCP threshold $\tau=0.5$, GeoDistill-Refine increases Image IoU from $0.5125\pm0.0560$ to $0.5491\pm0.0378$, Boundary F1 from $0.6863\pm0.1313$ to $0.7581\pm0.0841$, and recall from $0.8117\pm0.0343$ to $0.8826\pm0.0294$. The gains therefore persist without model-specific threshold calibration.

Deployment retains only the TinyUNet segmentation path. The student contains approximately 0.263 M parameters and 18.22 GFLOPs and requires about 1.1 ms per image on the RTX 4090. The six-prompt SAM 3 teacher contains 840.5 M parameters and requires 447.9 ms per image. Both the teacher and pseudo-geometry construction are confined to offline training.

\begin{table}[htbp]
  \centering
  \caption{Computational efficiency of the teacher and deployed student.}
  \label{tab:efficiency}
  \scriptsize
  \setlength{\tabcolsep}{2pt}
  \begin{tabularx}{\columnwidth}{@{}Zcccc@{}}
    \toprule
    Model & Parameters & \makecell{Latency\\(ms/image)} & \makecell{Throughput\\(image/s)} & \makecell{Peak memory\\(MB)} \\
    \midrule
    SAM 3 teacher& 840.5 M & 447.9 & 2.23 & 4234.0 \\
    TinyUNet student & 0.263 M & 1.1 & 900 & 102.7 \\
    \bottomrule
  \end{tabularx}
\end{table}

Plain Student and GeoDistill-Refine share the same deployed mask path. Table~\ref{tab:efficiency} excludes the training-only SDF and skeleton heads, and latency excludes disk input. The measurements characterize relative cost on the RTX 4090 and should not be interpreted as onboard-processor performance.

Figure~\ref{fig:qualitative-results} localizes the improvements summarized by the aggregate metrics. In rows (a)--(c), GeoDistill-Refine improves Image IoU by 0.114, 0.047, and 0.078 and Boundary F1 by 0.344, 0.057, and 0.202, respectively. The thin-structure example also raises small-region IoU from 0.286 to 0.929. Under the intense glare in row (d), however, Image IoU decreases from 0.228 to 0.177 and Boundary F1 from 0.905 to 0.742. Both students inherit the teacher-derived systematic foreground omission in this case, showing that pseudo-geometric refinement cannot recover structures absent from the pseudo-supervision.

\subsection{Ablation of Geometry Refinement}\label{sec:ablation}

\subsubsection{Effect of Silhouette-First Warm-Start Refinement}\label{sec:warm-start-ablation}

\begingroup\setlength{\emergencystretch}{1em}
Mechanistic ablations use random seed 42 on the SpaceSense-\allowbreak Bench KCP development set. Data, pseudo-labels, student architecture, and threshold calibration remain fixed. These experiments diagnose the training mechanism and do not replace the three-seed main results.
\par\endgroup

Joint training with the full geometric objective from scratch yields only 0.4509 Image IoU after 10 epochs. Extending it to the same 30-epoch budget produces 0.5645 Image IoU and 0.7876 Boundary F1, still below the two-stage result of 0.6056 and 0.8683. Training with only the SDF from scratch is also weaker than the Stage-I warm start: the former obtains 0.5214 and 0.7083, whereas the latter reaches 0.5979 and 0.8622. Under the present architecture and loss configuration, establishing the primary silhouette before introducing noisy pseudo-geometry is therefore more effective than joint optimization from random initialization. This observation is not generalized to other architectures or geometric objectives.

\subsubsection{Geometric Components and Reliability Gating}\label{sec:component-ablation}

Table~\ref{tab:component-ablation} compares the geometric components and reliability gate using the same Stage-I checkpoint, 10 refinement epochs, and threshold-calibration protocol.

Gated SDF is the strongest reduced configuration, identifying SDF as the principal signal in this setup. The full model adds 0.0077 Image IoU and 0.0061 Boundary F1, but area and skeleton show no monotonic standalone gain.Removing the gate lowers Image IoU from 0.6056 to 0.5917 and Boundary F1 from 0.8683 to 0.8588, supporting its role in limiting anomalous pseudo-geometry. The independent effects of the skeleton and area auxiliaries remain unresolved in this single-seed ablation.

\begin{table}[htbp]
  \centering
  \caption{Ablation of silhouette-first warm-start refinement on KCP.}
  \label{tab:warm-start-ablation}
  \scriptsize
  \setlength{\tabcolsep}{1.8pt}
  \begin{tabularx}{\columnwidth}{@{}Zccccc@{}}
    \toprule
    Training configuration & \makecell{Stage\\I} & \makecell{Stage\\II} & \makecell{Total\\epochs} & \makecell{Image\\IoU} & \makecell{Boundary\\F1} \\
    \midrule
    Plain Student & 20 & -- & 20 & 0.5697 & 0.8024 \\
    Full geometry, from scratch & -- & 10 & 10 & 0.4509 & 0.6329 \\
    Full geometry, from scratch & -- & 30 & 30 & 0.5645 & 0.7876 \\
    SDF only, from scratch & -- & 10 & 10 & 0.5214 & 0.7083 \\
    Stage I $\rightarrow$ gated SDF & 20 & 10 & 30 & 0.5979 & 0.8622 \\
    GeoDistill-Refine & 20 & 10 & 30 & \textbf{0.6056} & \textbf{0.8683} \\
    \bottomrule
  \end{tabularx}
\end{table}

\section{Conclusion}\label{sec:conclusion}

GeoDistill-Refine distills offline multi-prompt SAM 3 pseudo-masks into a compact spacecraft foreground segmenter without using manual masks for student updates. Its two-stage schedule first learns the foreground silhouette and then introduces reliability-gated SDF, skeleton, and area supervision. On the pre-reserved HJM lockbox set, this design improves Image IoU and Boundary F1 by 0.0456 and 0.1380 over Plain Student. The KCP ablations suggest that the silhouette-first warm start and gated SDF account for most of the observed refinement benefit under the current TinyUNet configuration. SPEED+ and TANGO provide complementary evidence on difficult laboratory imagery, where the quality-adaptive configuration improves boundary quality or foreground precision but does not exceed the external reproduction baseline on every metric.

\begin{table}[htbp]
  \centering
  \caption{Ablation of geometric components and reliability gating on KCP.}
  \label{tab:component-ablation}
  \scriptsize
  \setlength{\tabcolsep}{1.3pt}
  \begin{tabularx}{\columnwidth}{@{}Zcccccc@{}}
    \toprule
    Refinement configuration & SDF & Skel. & Area & Gate & IoU & BF1 \\
    \midrule
    Gated SDF only & $\checkmark$ &  &  & $\checkmark$ & 0.5979 & 0.8622 \\
    Gated SDF + skeleton & $\checkmark$ & $\checkmark$ &  & $\checkmark$ & 0.5487 & 0.7992 \\
    Gated SDF + area & $\checkmark$ &  & $\checkmark$ & $\checkmark$ & 0.5855 & 0.8437 \\
    SDF + skeleton + area, no gate & $\checkmark$ & $\checkmark$ & $\checkmark$ &  & 0.5917 & 0.8588 \\
    GeoDistill-Refine & $\checkmark$ & $\checkmark$ & $\checkmark$ & $\checkmark$ & \textbf{0.6056} & \textbf{0.8683} \\
    \bottomrule
  \end{tabularx}
\end{table}

The deployed model contains approximately 0.263 M parameters and requires about 1.1 ms per image on an RTX 4090; the teacher and auxiliary geometry heads are absent from inference. The evidence is limited to binary segmentation with validation masks used for checkpoint selection and threshold calibration. Only SpaceSense-Bench supports conclusions about unseen spacecraft identities, and the independent contributions of the skeleton and area objectives remain unresolved.

\printcredits

\bibliographystyle{cas-model2-names-unsorted}
\bibliography{references}

\end{document}